\documentclass[preprint,11pt]{elsarticle}

\usepackage{graphicx}
\usepackage{float}
\usepackage{amsmath}
\usepackage{booktabs}
\usepackage{algorithm}
\usepackage{algpseudocode}
\usepackage{hyperref}
\usepackage{xcolor}
\usepackage{float}
\usepackage{acronym}
\usepackage{tabularx}
\usepackage{multicol}
\usepackage{array}
\usepackage{makecell}
\usepackage{graphicx}
\usepackage{subcaption}

\usepackage{amssymb}

\journal{}

\usepackage[utf8]{inputenc}

\begin{document}

\begin{frontmatter}

\title{Automating Data-Driven Modeling and Analysis for Engineering Applications using Large Language Model Agents}

\author[inst1]{Yang Liu\corref{cor1}}
\ead{y-liu@tamu.edu}

\author[inst1]{Zaid Abulawi}

\author[inst1]{Abhiram Garimidi}

\author[inst1]{Doyeong Lim}

\affiliation[inst1]{organization={Department of Nuclear Engineering, Texas A\&M University},
            country={USA}}
            
\cortext[cor1]{Corresponding author}

\begin{abstract}
Modern engineering increasingly relies on vast datasets generated by experiments and simulations, driving a growing demand for efficient, reliable, and broadly applicable modeling strategies. There is also heightened interest in developing data-driven approaches, particularly neural network models, for effective prediction and analysis of scientific datasets. Traditional data-driven methods frequently involve extensive manual intervention, limiting their ability to scale effectively and generalize to diverse applications. In this study, we propose an innovative pipeline utilizing Large Language Model (LLM) agents to automate data-driven modeling and analysis, with a particular emphasis on regression tasks. We evaluate two LLM-agent frameworks: a multi-agent system featuring specialized collaborative agents, and a single-agent system based on the Reasoning and Acting (ReAct) paradigm. Both frameworks autonomously handle data preprocessing, neural network development, training, hyperparameter optimization, and uncertainty quantification (UQ). We validate our approach using a critical heat flux (CHF) prediction benchmark, involving approximately 25,000 experimental data points from the OECD/NEA benchmark dataset. Results indicate that our LLM-agent-developed model surpasses traditional CHF lookup tables and delivers predictive accuracy and UQ on par with state-of-the-art Bayesian optimized deep neural network models developed by human experts. These outcomes underscore the significant potential of LLM-based agents to automate complex engineering modeling tasks, greatly reducing human workload while meeting or exceeding existing standards of predictive performance.
\end{abstract}


\end{frontmatter}

\section{Introduction}
\label{sec:intro}
Modern science and engineering endeavors increasingly generate massive volumes of data through high-fidelity simulations and high-resolution experiments. Efficiently leveraging these extensive datasets for scientific insights and decision-making has spurred considerable interest in data-driven modeling using Artificial Intelligence (AI) and Machine Learning (ML) methods, particularly deep neural networks (DNNs). These methods excel at capturing intricate patterns within complex datasets, significantly enhancing predictive capabilities compared to traditional empirical approaches. In nuclear engineering modeling and simulation (M\&S), DNN-based data-driven model has achieved many successful applications. For instance, DNNs trained on interface-tracking data can reproduce boiling heat-transfer behavior with far greater speed than traditional closures~\cite{liu2018data}. In other work, convolutional-recurrent architectures emulate coarse-mesh turbulence for real-time transient analysis~\cite{liu2022data}, and encoder-decoder networks learn complex flow-field mappings in pebble-bed reactors~\cite{lanade2025high}. Furthermore, data-driven approaches have been used to create a hybrid porous-media model that extends CFD to full-core simulations~\cite{wang2021hybrid} and to develop the data-driven closure that integrates DNN model with traditional numerical solver for improved modeling capability~\cite{liu2022sam}, and to build whole-system digital twins that use graph neural networks as scalable surrogates for advanced reactors~\cite{Liu_gnn_dt}. 

Despite the promise of data-driven modeling, building a reliable model for each new engineering application remains a labor-intensive, case-by-case effort. The heterogeneity of engineering systems and conditions means no single modeling approach fits all scenarios. Engineers and data scientists must perform tasks like feature selection, data processing, model development, model training, hyperparameter tuning, and uncertainty analysis for each case, a process that can be time-consuming and requires substantial expertise. The lack of standardized frameworks or benchmarks in many engineering domains has further hampered wide adoption of advanced ML techniques. In nuclear engineering, for instance, the need for consistency and verification led the Organization for Economic Co-operation and Development (OECD)'s Nuclear Energy Agency (NEA) to establish dedicated benchmark exercises for AI/ML \cite{le2023benchmark}. This underscores both the demand and the current gap: while data-driven methods are recognized as valuable, the community lacks general, easy-to-use solutions to automate and streamline model development across different problems.

Large Language Models (LLMs) have recently emerged as powerful general-purpose reasoning engines that could help bridge this gap. State-of-the-art LLMs like OpenAI's o3-pro, Google's Gemini-2.5-pro, and Anthropic's Claude 4 have demonstrated impressive capabilities in coding, data analysis, and complex reasoning.  Recent works highlight a rapid convergence of LLMs with structured knowledge and agentic reasoning. A broad overview of this landscape is given by Yang \textit{et al.}, who survey techniques for integrating LLMs with knowledge-based methods \cite{yang2025comprehensive}. Several works push knowledge-graph understanding: Yang \textit{et al.} enhance zero-shot knowledge-graph completion via semantic cues \cite{yang2024enhancing}, Dai \textit{et al.} demonstrate that LLMs natively reason over graph structure \cite{dai2025large}, and Huang \textit{et al.} design knowledge-graph‐driven prompts for answering long-tail factual questions \cite{huang2025prompting}. Domain-focused studies embed expert knowledge into LLMs for programming-skill tracing \cite{sun2025harnessing}, emergency triage and multi-label medical diagnosis \cite{shen2025knowledge,xie2024knowledge}, causal traffic-risk assessment with chain-of-thought reasoning \cite{zhong2025large}, and soft prompt tuning to boost dense retrieval \cite{peng2025soft}. Schuerkamp \textit{et al.} further show that LLMs coupled with fuzzy cognitive maps can reconcile conflicting expert systems \cite{schuerkamp2025automatically}. Radaideh \textit{et al.} leverages LLM to perform large-scale sentiment analysis on internet to quantify social bias and sentiment \cite{radaideh2025fairness}. On the other hand, despite these promising demonstrations, leveraging LLM agents specifically for comprehensive data-driven scientific and engineering analyses remains relatively unexplored, presenting significant untapped potential for automating complex analytical workflows.

Notably, LLM-based agents, based on LLM and integrated with customized tools and libraries, can be designed to autonomously perform multi-step tasks by combining the LLM's reasoning with the ability to take actions such as running code or querying data sources \cite{sapkota2025ai}. Early studies have shown that LLM agents are effective across many application domains and are being adopted in computer science domains as productivity tools. One example is Ndum \textit{et al.} developed LLM agents to automate Monte Carlo simulations for nuclear engineering applications based on the FLUKA code \cite{NDUM2025100555}. However, their use for scientific data analysis and engineering problems is still limited and poses unique challenges. Data-driven modeling workflows for scientific problems often involve long-term, interconnected tasks that require domain knowledge and dynamic adaptation to intermediate results \cite{Ni2024MechAgents:Knowledge}. Naïvely applying an LLM to such problems tends to yield only basic analyses, as the model might rely on its internal knowledge and simple heuristics. Indeed, a recent benchmark BLADE suite \cite{gu-etal-2024-blade} found that while LLMs possess considerable general knowledge, they often default to superficial analysis and hallucinations, whereas giving an LLM the ability to interact with real data (through tools or code) leads to more accurate analytical decisions \cite{sapkota2025ai}. These findings highlight the importance of an agentic approach, where an LLM actively engages with data and tools, rather than a static question-answer approach.

Among different LLM agent approaches, two promising paradigms have emerged in this context. The first is the ReAct framework (Reasoning and Acting) for single-agent prompting. In the ReAct paradigm, the LLM interleaves its chain-of-thought reasoning with tool-using actions in an interactive loop \cite{yao2023react}. This allows the model to plan a solution step-by-step, execute computations (e.g. running code, retrieving information), observe the results, and adjust its strategy. The ReAct approach has been shown to improve task performance over methods that rely on reasoning-only or acting-only, by synergizing the two. The second paradigm involves multi-agent systems: instead of a single monolithic agent, multiple specialized LLM agents can be organized to collaborate on sub-tasks. For example, a recent framework MetaGPT encodes human Standard Operating Procedures into prompts and assigns different expert roles to multiple LLM agents, who then work in an ``assembly line" to tackle complex problems collaboratively \cite{hong2024metagpt}. Such multi-agent collaborations can reduce errors (by having agents verify each other's outputs) and handle complexity via divide-and-conquer \cite{Topsakal2023CreatingFast, wu2023autogen}. Overall, these developments suggest that LLM agents, either a well-instructed solo agent or a coordinated team of agents, could automate significant parts of the data-driven modeling workflow in engineering. However, to date there have been few demonstrations of these techniques applied to real scientific/engineering modeling tasks, especially using the latest-generation models.

To provide a robust evaluation of these advanced agentic systems, we leverage the framework established by the OECD/NEA's international benchmark on AI/ML. The inaugural phase of this benchmark directly addresses the limitations of traditional engineering models by focusing on the prediction of critical heat flux (CHF), a thermal-hydraulic limit of paramount importance to nuclear reactor safety. CHF marks the point where heat transfer from the fuel rod surface degrades significantly, potentially causing rapid cladding temperature increases and fuel damage. While traditional methods like the 2006 Groeneveld look-up table (LUT) have been foundational, they possess known uncertainties stemming from the limited scope of their underlying data \cite{groeneveld20072006}. The NEA benchmark therefore presents a well-defined challenge: to develop superior, data-driven models by leveraging the comprehensive US Nuclear Regulatory Commission (NRC) CHF database, which contains over 24,000 experimental points. Crucially, the benchmark ensures rigorous validation by requiring participants to evaluate their models against a separate, blind test dataset, providing an ideal and standardized testbed for assessing the capabilities of our proposed automated modeling pipeline.

In this work, we directly address this benchmark challenge by developing and comparing two distinct LLM-based agentic systems: a ReAct-style single agent and a collaborative multi-agent team. We task these systems with autonomously performing the end-to-end data-driven analysis required by the NEA's CHF benchmark. Our objective is to demonstrate that, with minimal human intervention, these agents can successfully navigate the entire modeling workflow including data exploration, feature selection, data processing, model development, training, performance evaluation, and UQ. By subjecting our automated agents to this rigorous, real-world engineering problem, we aim to critically assess their potential to accelerate and democratize complex data-driven modeling in demanding scientific domains.

The remainder of this paper is organized as follows. In Section \ref{sec:methodology}, we detail our methodology, describing the design and implementation of both the multi-agent and ReAct agent systems for data-driven modeling; Section \ref{sec:case_study} introduces the CHF case study based on the OECD/NEA benchmark; Section \ref{sec:results} discusses the results of LLM agents and compare it with human-developed baseline model and existing empirical CHF lookup table; finally Section \ref{sec:conclusions} concludes with an analysis of the success and limitations of the current work and outline directions for future improvements.

\section{Methodology}
\label{sec:methodology}

\begin{figure}[H]
    \centering
    \includegraphics[width=1.0\textwidth]{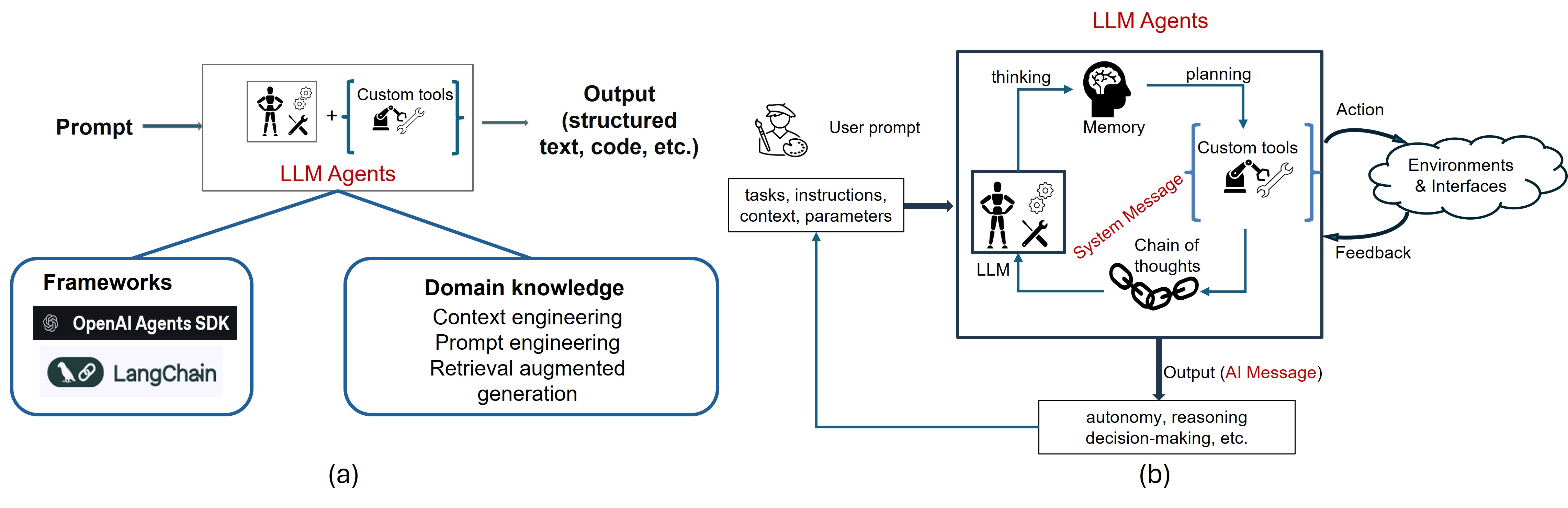}
    \caption{(a) An agentic system leveraging a pre-trained LLM, supporting libraries, and domain knowledge. (b) A typical LLM agent workflow, including environment interaction, feedback processing, internal reasoning, and tool execution to achieve specified goals.}
    \label{fig:llm_agents}
\end{figure}

To implement and evaluate these agentic LLM systems, we constructed a robust pipeline utilizing state-of-the-art tools, as depicted in Figure \ref{fig:llm_agents}. Our framework is built upon the OpenAI's Agent SDK \cite{openai-agent}, which provides the core functionalities for defining agents, managing context, and orchestrating tool use. The agents were equipped with a sandboxed Python interpreter, giving them access to a rich scientific computing stack, including PyTorch \cite{paszke2019pytorch} for building and training and evaluating DNNs. The code and configuration for our agent systems are made publicly available in a GitHub repository to facilitate reproducibility. In this work, we used OpenAI's API to access its pre-trained LLMs including GPT-4.1 and O3, but the developed framework and workflow can be applied to any types of LLMs, including locally deployed open-source LLM.

\subsection{Deep Neural Network and Deep Ensemble}

To provide the agents with a powerful and flexible modeling tool, we focus on DNNs, which has good ability to learn complex, non-linear relationships directly from data. At the core of our approach is the Multi-Layer Perceptron (MLP), a foundational class of feedforward DNN. An MLP is a universal function approximator, meaning that with sufficient complexity, it can learn to map any continuous function \cite{hornik1989multilayer}. This makes it an exceptionally versatile tool for a wide range of engineering regression and classification tasks. 

Let $\mathbf{x} \in \mathbb{R}^D$ be an input vector with $D$ features. A DNN model, specifically a multi-layer perceptron (MLP), is constructed with $L$ hidden layers. The transformation at each layer $l \in \{1, \dots, L\}$ is given by:

\begin{equation}
\mathbf{h}_l = \sigma(\mathbf{W}_l \mathbf{h}_{l-1} + \mathbf{b}_l),
\end{equation}

where $\mathbf{h}_0 = \mathbf{x}$, $\mathbf{W}_l$ is the weight matrix, $\mathbf{b}_l$ is the bias vector for layer $l$, and $\sigma$ is a non-linear activation function (e.g., ReLU). The output layer produces the prediction. For a direct regression, the output $\hat{y} \in \mathbb{R}$ is:

\begin{equation}
\hat{y} = \mathbf{w}_{out}^T \mathbf{h}_L + b_{out}.
\end{equation}

A key requirement for engineering applications is not just an accurate prediction, but also a reliable estimate of its uncertainty. UQ is an active research topics in engineering modeling and risk analysis. For example, recent work includes developing modular Bayesian approaches to rigorously propagate epistemic and aleatory uncertainty in two-phase flow and boiling predictions~\cite{liu2019uncertainty}; using machine-learning–assisted Bayesian calibration to refine multiphase-CFD models of bubbly flows~\cite{liu2021uncertainty} and their closure relations~\cite{liu2023uncertainty}; elucidating measurement error in experimental data through ray-optics-based analyses of particle-image velocimetry in bubbly flows~\cite{liu2020uncertainty}; and performing benchmark studies to quantify model-form and parametric uncertainty in system-level transients~\cite{liu2023benchmarking}.

The uncertainty of DNN predictions can be decomposed into two fundamental types: \textit{aleatory uncertainty} captures the inherent noise or statistical variability in the data, and \textit{epistemic uncertainty} reflects the model's lack of confidence due to limited training data. To provide a comprehensive and trustworthy assessment, both sources of uncertainty must be quantified. For this purpose, we employ the deep ensemble method \cite{lakshminarayanan2017simple}, which has been demonstrated to be a simple and scalable appraoch for the UQ of DNNs \cite{ABULAWI2025111353, liu2021uncertainty_a}. The core idea is to train an ensemble of several identical DNNs, each with a different random weight initialization. The disagreement in the predictions across these models serves as a natural measure of epistemic uncertainty, as models will diverge more in regions of the input space where data is sparse. Simultaneously, each individual network in the ensemble is trained to capture aleatory uncertainty by predicting the parameters of a probability distribution for the target variable. Assuming the target $y$ follows a Gaussian distribution conditioned on the input $\mathbf{x}$, such that $p(y|\mathbf{x}) = \mathcal{N}(y | \mu(\mathbf{x}), \sigma^2(\mathbf{x}))$, each network is trained by minimizing the negative log-likelihood loss:

\begin{equation}
\mathcal{L}(\theta) = \frac{1}{N} \sum_{i=1}^N \left[ \frac{(y_i - \mu(\mathbf{x}_i; \theta))^2}{2\sigma^2(\mathbf{x}_i; \theta)} + \frac{1}{2}\log\sigma^2(\mathbf{x}_i; \theta) \right]
\end{equation}

where $\theta$ represents the network parameters. This dual approach allows the ensemble as a whole to decompose the total predictive uncertainty into its aleatory and epistemic components.

In a deep ensemble, $M$ identical networks are trained independently, each with a different setup of hyperparameters and random initialization of its weights $\theta_m$. Given an input $\mathbf{x}$, each model $m$ produces a predictive distribution $p(y|\mathbf{x}, \theta_m) = \mathcal{N}(y|\mu_m(\mathbf{x}), \sigma_m^2(\mathbf{x}))$. The ensemble's final predictive distribution is approximated by an equally weighted mixture of these individual Gaussian distributions:

\begin{equation}
p(y|\mathbf{x}, \{\theta_m\}_{m=1}^M) = \frac{1}{M} \sum_{m=1}^M \mathcal{N}(y|\mu_m(\mathbf{x}), \sigma_m^2(\mathbf{x})).
\end{equation}

The final predictive mean $\bar{\mu}(\mathbf{x})$ and total variance $\sigma^2_{\text{total}}(\mathbf{x})$ are calculated using the law of total variance, which provides an elegant decomposition of the two types of uncertainty:

\begin{equation}
\bar{\mu}(\mathbf{x}) = \frac{1}{M} \sum_{m=1}^M \mu_m(\mathbf{x}),
\end{equation}

\begin{equation}
  \sigma^2_{\text{total}}(\mathbf{x}) = \underbrace{\frac{1}{M} \sum_{m=1}^M \sigma_m^2(\mathbf{x})}_{\text{aleatory}} + \underbrace{\frac{1}{M} \sum_{m=1}^M (\mu_m(\mathbf{x}) - \bar{\mu}(\mathbf{x}))^2}_{\text{Epistemic}}.  
\end{equation}

The agents in our pipeline can be tasked with constructing, training, and evaluating such deep ensemble models to provide both accurate predictions and robust uncertainty estimates, which is critical for trustworthy deployment in engineering analysis.

\subsection{Multi-Agent System}

\begin{figure}[H]
    \centering
    \includegraphics[width=1.0\textwidth]{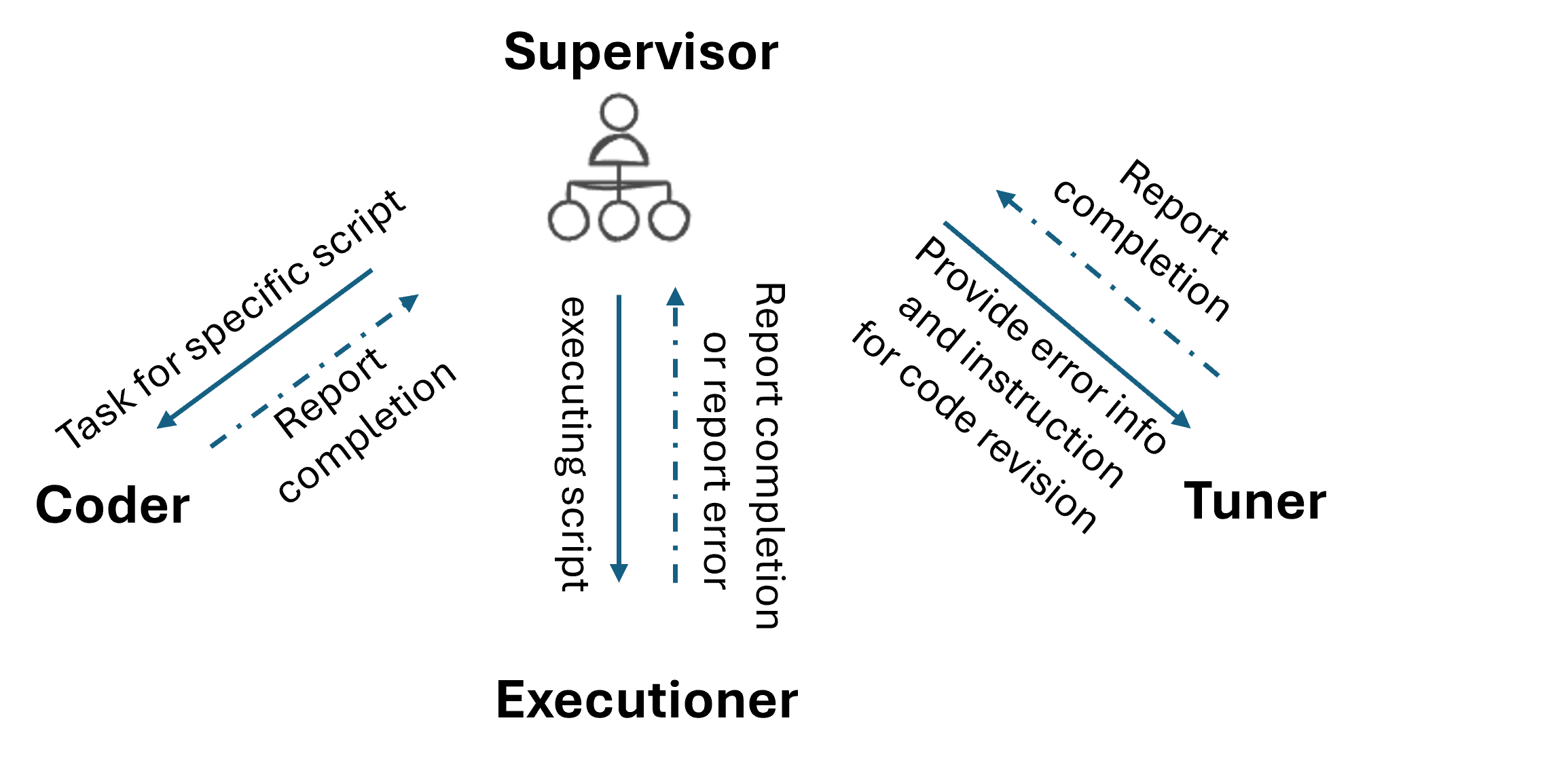}
    \caption{Supervisor-centric architecture of the multi-agent system, illustrating the interaction between the supervisor and the coder, executioner, and tuner agents.}
    \label{fig:multi_agent}
\end{figure}

The multi-agent system is designed as a hierarchical framework where a complex problem is decomposed into simpler sub-tasks, each handled by a specialized agent. To ensure consistent context and efficient information exchange, we implemented a supervisor-centric, or hub-and-spoke, architecture. As depicted in Figure~\ref{fig:multi_agent}, a central Supervisor Agent acts as the sole orchestrator, directly managing and communicating with all other specialized agents. The subordinate agents do not communicate with each other; instead, all task delegations, results, and error logs are routed through the Supervisor. This design prevents information silos, simplifies the workflow, and provides a single point of control for state management and error handling.

We define a set of agents with distinct, non-overlapping responsibilities, detailed in Table~\ref{tab:multi_agent_workflow}:

\begin{table}[H]
\centering
\caption{Overview of the key agents in the proposed multi-agent system.}
\label{tab:multi_agent_workflow}
\renewcommand{\arraystretch}{1.05} 
\small 
\begin{tabularx}{\textwidth}{
  >{\bfseries\raggedright\arraybackslash}p{2.8cm} 
  >{\raggedright\arraybackslash}p{3.5cm} 
  X
}
\toprule
\textbf{Role/Component} & \textbf{Core Implementation} & \textbf{Description} \\
\midrule
Supervisor Agent & \textit{Supervisor} class (O3 model) & Orchestrates the workflow, manages state and data context, delegates tasks, and synthesizes the final report. \\
Coding Agent & Specialized Agent (GPT-4.1) & Generates Python scripts for model architecture, training, and evaluation upon request from the Supervisor. \\
Tuning Agent & Specialized Agent (GPT-4.1) & Receives a failed script and error log from the Supervisor, and returns a corrected version of the script. \\
Execution Agent & \textit{execute\_script} function & Runs generated scripts in a sandboxed environment and captures outputs for the Supervisor. \\
Data Management & \textit{ProjectContext} class & Centralized, consistent source of truth for all data and artifact file paths, managed by the Supervisor. \\
\bottomrule
\end{tabularx}
\end{table}

The workflow, illustrated in Algorithm~\ref{alg:multi_agent}, is a stateful process managed entirely by the Supervisor. It begins by invoking the Coding Agent to generate the necessary scripts (model definition, training, and evaluation). For each generated script, the Supervisor calls upon the Execution Agent to run it. If the execution fails, the Supervisor passes the script and the error log to the Tuning Agent for correction, initiating a self-correction loop that continues until the script runs successfully. This generate-execute-tune cycle proceeds until the entire pipeline is completed and the final model and its evaluation are produced.

\begin{algorithm}[H]
\caption{Multi-Agent System Workflow}
\label{alg:multi_agent}
\begin{algorithmic}[1]
\State \textbf{Initialize:} Supervisor, Coding, Tuning, Execution Agents, ProjectContext
\State \textbf{Input:} High-level task description $T$, Dataset $D$
\State $script\_path, status \leftarrow \textit{generate\_and\_execute('model')}$
\While{$status$ is Error}
    \State \textit{Supervisor.tune\_script($script\_path, error\_log$)}
    \State $status \leftarrow \textit{execute\_script(script\_path)}$
\EndWhile
\State $script\_path, status \leftarrow \textit{generate\_and\_execute('train')}$
\While{$status$ is Error}
    \State \textit{tune\_script($script\_path, error\_log$)}
    \State $status \leftarrow \textit{execute\_script(script\_path)}$
\EndWhile
\State $script\_path, status \leftarrow \textit{generate\_and\_execute('evaluate')}$
\While{$status$ is Error}
    \State \textit{tune\_script($script\_path, error\_log$)}
    \State $status \leftarrow \textit{execute\_script(script\_path)}$
\EndWhile
\State \textit{Synthesize\_Final\_Report()}
\State \textbf{Output:} Final Model and Performance Report
\end{algorithmic}
\end{algorithm}

A key feature of this architecture is the dynamic generation of prompts for the Coding and Tuning Agents. Instead of relying on static prompts, the Supervisor constructs them at runtime using f-strings, populating them with precise, absolute file paths retrieved from a central \textit{ProjectContext} object. This ensures that the generated code is always tailored to the specific environment of the current run, greatly enhancing reliability and reproducibility. This dynamic nature is also a powerful tool for debugging, as the exact prompts sent to the LLMs can be logged and inspected, providing a clear audit trail of the agent's instructions.

To ensure accurate and efficient context exchange, the system relies on two primary mechanisms. First, the \textit{ProjectContext} object is instantiated at the start of the workflow and managed by the Supervisor. This object centralizes all critical path information, eliminating ambiguity and ensuring that every agent and every generated script operates on the correct files. Second, the Supervisor maintains the high-level state of the workflow in a simple JSON file. This state persistence allows the system to be stopped and resumed, and it provides a clear record of which stages have been completed successfully. This structured approach to context and state management is crucial for the system's robustness and scalability.

\subsection{Single ReAct-Agent System}

\begin{figure}[H]
    \centering
    \includegraphics[width=1.0\textwidth]{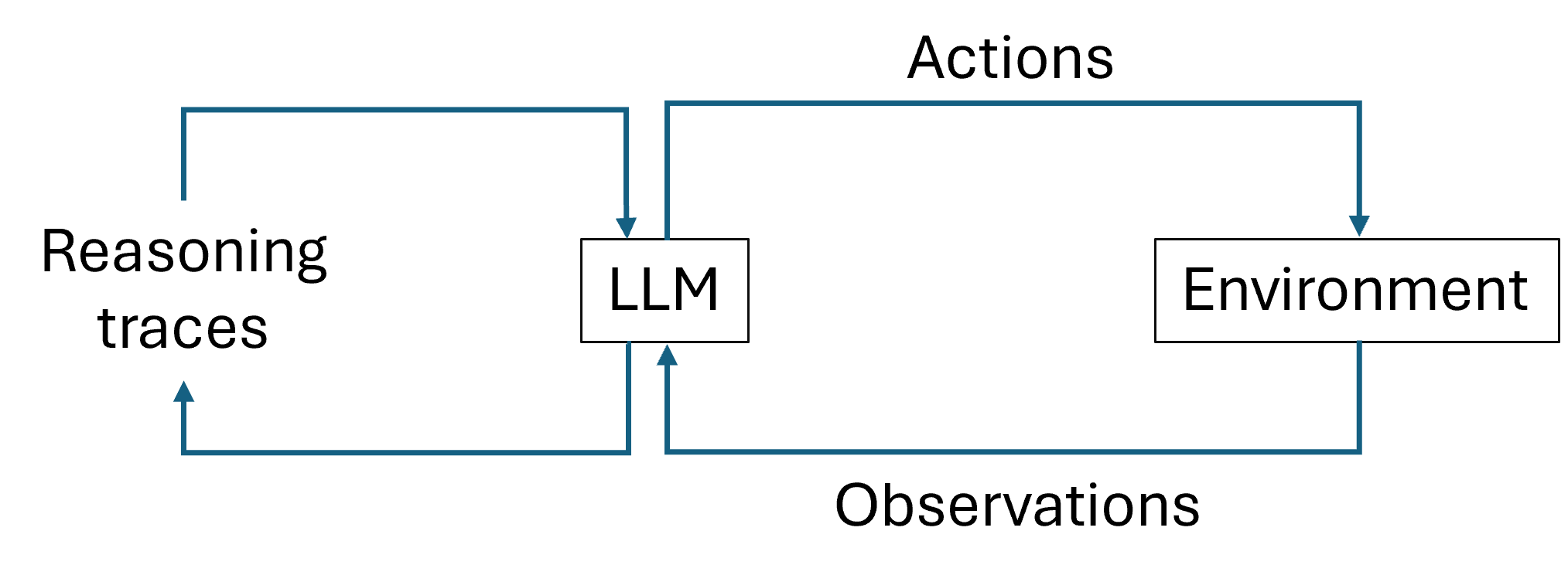}
    \caption{The ReAct-based architecture of the single-agent system, illustrating the iterative loop between reasoning traces, actions, and observations\cite{yao2023react}.}
    \label{fig:react}
\end{figure}

The single-agent system is based on the ReAct paradigm \cite{yao2023react}, a powerful prompting strategy that enables an LLM to tackle complex tasks by synergizing its internal reasoning capabilities with external tool use. In this framework, a single, versatile agent is responsible for the entire modeling workflow, from data analysis to final evaluation. The core of the ReAct method is an iterative loop where the agent interleaves thoughts, actions, and observations, as depicted in Figure \ref{fig:react}.

Given a task, the agent first generates a ``Thought," which is a textual chain-of-thought reasoning step to break down the problem and devise a plan. Based on this thought, it then formulates an ``Action," which is typically a call to an external tool (e.g., executing a Python code snippet). The output from the tool is then fed back to the agent as an ``Observation". This observation informs the agent's next thought, allowing it to assess its progress, analyze results, and dynamically adjust its strategy. This process can be formalized as follows: at each step $t$, the agent has access to a history of previous steps, $H_{t-1} = \{(T_1, A_1, O_1), \dots, (T_{t-1}, A_{t-1}, O_{t-1})\}$. The agent then generates a new thought and action pair:
\[ T_t, A_t = \text{LLM}(\text{Task}, H_{t-1}) \]
The action $A_t$ is executed by a tool, yielding an observation $O_t = \text{Tool}(A_t)$. This new triple $(T_t, A_t, O_t)$ is appended to the history for the next step.

Our implementation of this paradigm is encapsulated within the \textit{CHFReActAgent} class, which orchestrates the entire modeling workflow, from data ingestion to UQ. The process begins when the \textit{run} method is called, which initializes a state manager and starts an asynchronous ReAct loop. A \textit{StateManager} object tracks the overall progress, persisting the current state (e.g., which steps like \textit{model\_generation} or \textit{training\_execution} are complete) to a JSON file. This ensures the agent's work can be resumed and provides a clear audit trail. Short-term memory is maintained as a list of recent thought-action-observation tuples, which are fed back into the reasoning prompt to provide immediate context for the next decision.

The agent's decision-making process, summarized in Table~\ref{tab:react_agent_workflow}, occurs in the \textit{\_think} method. Here, a detailed prompt is dynamically constructed, synthesizing the high-level task description with the current state, a list of available tools, critical file paths, and the short-term memory. The LLM's response constitutes the ``Thought''---a high-level plan for the next action. This plan is then passed to the \textit{\_act} method, which serves as a dispatcher. It parses the plan to determine which of its available tools should be called. Rather than a single, general-purpose Python interpreter, our agent is equipped with a discrete set of more robust, specialized tools, such as \textit{generate\_model}, \textit{execute\_python\_script}, and \textit{generate\_evaluation\_script}. This design choice enhances reliability by constraining the agent's actions to a set of well-defined operations.

\begin{table}[H]
\centering
\caption{Phases, core methods, descriptions, and key inputs/outputs of the single ReAct agent implementation.}
\label{tab:react_agent_workflow}
\begin{tabular}{
    >{\bfseries\raggedright\arraybackslash}p{2cm}
    >{\raggedright\arraybackslash}p{3.5cm}
    >{\raggedright\arraybackslash}p{4.5cm}
    >{\raggedright\arraybackslash}p{4.5cm}
}
\toprule
\textbf{Phase} & \textbf{Core Method(s)} & \textbf{Description} & \textbf{Key Inputs \& Outputs} \\
\midrule
Thought & \textit{\_think} & Constructs a dynamic prompt based on the current state and asks an LLM to generate a plan for the next step. & \textbf{In:} Current state, task description, memory. \newline \textbf{Out:} High-level action plan (text). \\
\addlinespace
Action & \textit{\_act}, tool functions & Parses the action plan and dispatches to a specific, predefined tool (e.g., \textit{generate\_model}, \textit{execute\_python\_script}). & \textbf{In:} Action plan (text). \newline \textbf{Out:} Dictionary with success status, logs, and paths to generated artifacts. \\
\addlinespace
Observation & \textit{\_observe} & Summarizes the outcome of the action into a concise observation message, which is added to the agent's memory. & \textbf{In:} Action result dictionary. \newline \textbf{Out:} Observation summary (text). \\
\bottomrule
\end{tabular}
\end{table}

Finally, the result of the action is processed in the \textit{\_observe} method. This step synthesizes a concise summary of the outcome (e.g., ``Model generated and saved successfully'' or ``Script execution failed with error...''). This ``Observation'' completes the loop and is added to the memory, informing the subsequent ``Thought'' step. A key feature of this approach is its capacity for self-correction. If an action results in an error, the error message is returned as part of the ``Observation''. The agent can then reason about the cause of the error in its next ``Thought'' and generate a corrected action, making the process highly resilient. This iterative cycle continues until all stages of the modeling pipeline are complete, as outlined in Algorithm~\ref{alg:react_agent}.

\begin{algorithm}[H]
\caption{Single-Agent (ReAct) System Workflow}
\label{alg:react_agent}
\begin{algorithmic}[1]
\State \textbf{Initialize:} ReAct Agent with Python tool access
\State \textbf{Input:} High-level task description $T$, Dataset $D$
\State History $H \leftarrow \emptyset$
\State Status $\leftarrow \text{InProgress}$
\While{Status $\neq$ \text{Complete}}
    \State Thought, Action $\leftarrow$ Agent.generate\_plan($T, H$)
    \If{Action is \text{finish\_task}}
        \State Status $\leftarrow \text{Complete}$
        \State \textbf{break}
    \EndIf
    \State Observation $\leftarrow$ PythonTool.execute(Action)
    \State $H \leftarrow H \cup \{(\text{Thought, Action, Observation})\}$
\EndWhile
\State \textbf{Output:} Final Model and Summary synthesized from $H$
\end{algorithmic}
\end{algorithm}

\section{Case Study: Critical Heat Flux Prediction}
\label{sec:case_study}

\subsection{OECD/NEA Critical Heat Flux Benchmark}

To demonstrate and evaluate the capabilities of our LLM-agent systems, we selected a challenging and safety-critical problem from nuclear reactor thermal-hydraulics: the prediction of CHF. This problem was chosen as the inaugural exercise for the OECD/NEA's international AI/ML benchmark effort \cite{le2023benchmark}. Our work, therefore, directly engages with a standardized problem defined by domain experts, providing a rigorous and relevant testbed for assessing agent performance.

The benchmark is centered on the US NRC CHF database, which is the largest publicly available dataset of its kind and was the basis for the well-known 2006 CHF look-up table. This comprehensive collection contains approximately 24,579 data points compiled from 59 separate experimental sources, ensuring a wide diversity of conditions suitable for developing robust models. The dataset provides measured geometrical and operational parameters, including the tube diameter ($D$), heated length ($L$), system pressure ($P$), and mass flux ($G$). It also includes calculated parameters derived from these measurements, most notably the local thermodynamic equilibrium quality ($X$). As shown in Table~\ref{tab:chf_variables}, these experiments cover an extensive range of operating conditions, which presents a significant learning challenge. The target variable for the regression task is the CHF ($q_{\text{CHF}}$) itself.

\begin{table*}[t]
\centering
\caption{Summary of Variable Ranges for CHF Dataset}
\label{tab:chf_variables}
\begin{tabular}{lcccccc}
\toprule
\textbf{Variable} & \textbf{CHF} [kW/m\textsuperscript{2}] & \textbf{P} [kPa] & \textbf{G} [kg/m\textsuperscript{2}/s] & \textbf{X} [-] & \textbf{D} [m] & \textbf{L} [m] \\
\midrule
Min Value & 50 & 100 & 8.2 & -0.497 & $2 \times 10^{-3}$ & 0.05 \\
Max Value & 16339.3 & 20000 & 7964 & 0.999 & $16 \times 10^{-3}$ & 20 \\
\bottomrule
\end{tabular}
\end{table*}

The problem was presented to our LLM agent systems as a high-level objective. The agents were only given the raw dataset and tasked to develop a DNN-based deep ensemble to predict CHF with UQ. This granted the agents full autonomy to formulate their own technical approach, tasking them with all steps from data exploration and feature selection to model training, evaluation, and UQ, thereby testing their end-to-end problem-solving abilities.

The evaluation of the agent-developed models follows the rigorous framework set by the NEA benchmark, which is designed to ensure scientific validity and prevent overfitting. First, model performance is compared against two primary benchmarks: the industry-standard 2006 Groeneveld CHF LUT, which serves as a baseline to be surpassed, and other state-of-the-art, human-developed deep ensemble based on Bayesian optimization as detailed in Section 2.4. Second, the agents' final models are subjected to an independent blind test. As part of the benchmark, the organizers provide a separate, unseen dataset for which the CHF values are unknown to the participants (and thus to the agents). This test assesses the true generalization capability of the models on data not used during training or tuning. The blind data, which includes ten specific ``slice datasets,'' as defined by the benchmark and detailed in Table~\ref{tab:slice_datasets}. Each slice varies a single input parameter (e.g., pressure or diameter) across its operational range while holding all other parameters nearly constant. This allows for a qualitative assessment of the model's physical consistency and its ability to capture known trends without overfitting to the training data. This comprehensive case study provides a robust test of our LLM agent frameworks on a real-world engineering problem, assessing their ability to automate complex modeling workflows and deliver scientifically valid results.

\begin{table}[H]
\centering
\caption{Summary of slice datasets used as “blind” testing cases for model performance evaluation.}
\label{tab:slice_datasets}
\begin{tabular}{lccccc}
\toprule
\textbf{Slice \#} & $\mathbf{D}$ [mm] & $\mathbf{L}$ [m] & $\mathbf{P}$ [kPa] & $\mathbf{G}$ [kg/m$^2$/s] & $\mathbf{X}$ [-] \\
\midrule
1 & 8.01 & \textbf{0 -- 20} & 9806 & 1000.0 & 0.587 \\
2 & 8.11 & \textbf{0 -- 20} & 2009 & 752.2 & 0.756 \\
3 & 8.00 & 0.998 & \textbf{0 -- 20000} & 2006.0 & 0.140 \\
4 & 13.40 & 3.658 & \textbf{0 -- 20000} & 2040.2 & 0.378 \\
5 & 8.14 & 1.943 & 9831 & 1519.5 & \textbf{-0.5 -- 1.0} \\
6 & \textbf{0 -- 16} & 6.000 & 9807 & 1003.3 & 0.529 \\
7 & 8.00 & 1.570 & 12750 & \textbf{0 -- 8000} & 0.144 \\
8 & 10.00 & 4.966 & 16000 & \textbf{0 -- 8000} & 0.343 \\
\bottomrule
\end{tabular}
\end{table}

\subsection{Baseline Model Using Bayesian Optimized Deep Ensemble}
\label{subsec:bo_baseline}

To establish a strong performance benchmark, we developed a baseline model using a deep ensemble where the hyperparameters of each constituent network were systematically tuned via Bayesian Optimization (BO)\cite{ABULAWI2025111353, Abulawi_2025}. This approach aims to find optimal model configurations that minimize prediction error on unseen data. The optimization was conducted using the `Ax' platform \cite{chang2019bayesian}, a state-of-the-art library for managing BO experiments.

The optimization objective was to minimize the Root Mean Square Error (RMSE) on a dedicated validation set.
\begin{equation}
\text{RMSE} = \sqrt{\frac{1}{n} \sum_{i=1}^{n} (y_i - \mu_i)^2}
\label{eq:rmse_bo}
\end{equation}
Here, $y_i$ is the true value, $\mu_i$ is the model's predicted mean for the $i$-th sample, and $n$ is the number of samples in the validation set. The dataset was partitioned into training (72\%), validation (18\%), and test (10\%) subsets. The BO process tuned seven key hyperparameters across the search space detailed in Table~\ref{tab:bo_hyperparameters}.

\begin{table}[H]
\centering
\caption{Hyperparameter search space for Bayesian Optimization.}
\label{tab:bo_hyperparameters}
\begin{tabular}{ll}
\toprule
\textbf{Hyperparameter} & \textbf{Search Space} \\
\midrule
Learning Rate & $[10^{-4}, 10^{-2}]$ \\
Weight Decay & $[10^{-4}, 10^{-2}]$ \\
Dropout Rate & $[0, 0.3]$ \\
Batch Size & $\{128, 256, 512\}$ \\
Hidden Layers & $\{6, 7\}$ \\
Hidden Units & $\{8, 16, 24, 32, 48, 64, 96\}$ \\
Activation Function & \{ReLU, LeakyReLU, GeLU, SeLU, ELU, Softplus\} \\
\bottomrule
\end{tabular}
\end{table}

Our optimization strategy involved a two-stage process designed for efficiency and robustness. First, we initialized the search with a Sobol sequence, which quasi-randomly samples the hyperparameter space to provide a diverse set of initial configurations. These initial points were used to build a reliable prior for the subsequent BO stage. Following initialization, the main BO loop began, iteratively refining the search. In each iteration, a Gaussian Process (GP) surrogate model was updated to approximate the objective function's landscape. We choose a noisy expected improvement acquisition function then guided the selection of the next hyperparameters to evaluate, effectively balancing the exploration of new regions with the exploitation of known high-performing ones. More details can be found in \cite{Abulawi_2025, ABULAWI2025111353}.

To foster model diversity and create a robust final ensemble, we executed five independent BO runs in parallel, each with a different random seed. The final deep ensemble was constructed by selecting the top 15 networks from the combined results of all five runs. This ensures the final model is not only accurate but also benefits from the diversity captured across different optimization trajectories.

\section{Results and Discussion}
\label{sec:results}

\subsection{Workflow Analysis of LLM Agents}

An essential requirement for LLM-driven agents is their ability to autonomously manage complex, multi-step data analysis and modeling workflows, including effectively detecting, diagnosing, and recovering from errors. Given the inherent stochastic nature of LLMs as probabilistic models, agents are expected to exhibit variations in their behavior even when repeatedly tasked with identical tasks. Therefore, assessing their robustness in successfully completing assigned tasks becomes a critical metric for evaluating overall agent performance. To quantitatively examine the robustness, reliability, and computational efficiency of the two developed agent systems, we conducted 10 independent trials for each approach. In each trial, agents were tasked to autonomously complete the entire workflow from data ingestion through to model evaluation, while key performance metrics were monitored and recorded.

Table~\ref{tab:agent_comparison_transposed} summarizes these trials. Both systems achieved comparable average RMSE, with the Multi-Agent System marginally outperforming the ReAct Single Agent in terms of average and minimum RMSE. However, the Multi-Agent System also exhibited a higher maximum RMSE, indicating greater variability and hence slightly lower consistency at its worst.

\begin{table}[h!]
\centering
\caption{Performance comparison between the Multi-Agent System and the ReAct Single-Agent System, based on CHF RMSE statistics, completion success rates, and average token usage on testing data.}
\label{tab:agent_comparison_transposed}
\begin{tabular}{lcc}
\toprule
\textbf{Metric} & \textbf{Multi-Agent System} & \textbf{ReAct Single Agent} \\
\midrule
Average CHF RMSE on testing data & 250.4 & 251.1 \\
Minimum CHF RMSE on testing data & 230.2 & 234.4 \\
Maximum CHF RMSE on testing data & 301.9 & 293.4 \\
Completed without error & 7 & 6 \\
Completed with one error & 3 & 3 \\
Completed with $\geq 2$ error & 0 & 1 \\
Fail to complete & 0 & 0 \\
Average token usage & 11,287 & 35,311 \\
\bottomrule
\end{tabular}
\end{table}

Regarding task completion robustness, both agent systems completed all 10 trials successfully. While the Multi-Agent System exhibited better reliability, completing 7 out of 10 trials without any errors, whereas the ReAct Single Agent achieved 6 error-free completions. Both systems completed 3 trials with exactly one error. However, the ReAct Single Agent encountered at least two errors in one trial, indicating its higher susceptibility to repeated failures within the same task cycle. A critical differentiating factor between the two systems was computational efficiency, measured by average token consumption. The Multi-Agent System demonstrated better efficiency, utilizing approximately 68\% fewer tokens on average (11,287 tokens per trial) compared to the ReAct Single Agent (35,311 tokens per trial). This considerable difference highlights the advantage of task specialization across multiple agents, reducing redundancy and computational overhead associated with iterative single-agent reasoning.

\begin{figure}[H]
    \includegraphics[width=1.0\textwidth]{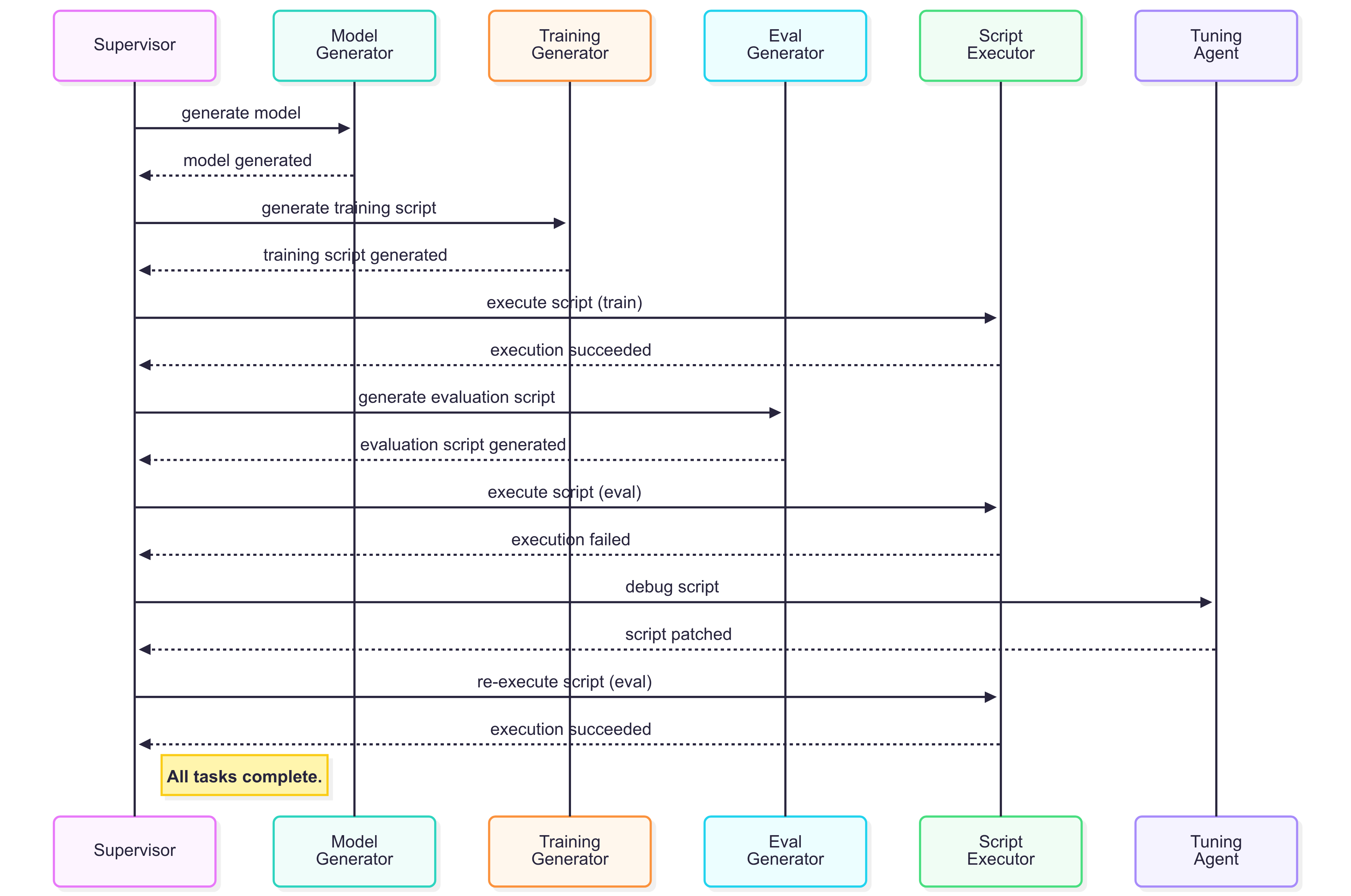}
    \caption{Workflow of the multi-agent system illustrating the handling of a code execution failure, and successful re-execution.}
    \label{fig:multiagent_flow}
\end{figure}

Furthermore, although both systems effectively completed the tasks, their operational workflows revealed distinct problem-solving strategies and error-correction mechanisms. The multi-agent system executed the workflow through a structured, top-down decomposition of tasks managed by the central Supervisor Agent. A typical run, as illustrated in Figure~\ref{fig:multiagent_flow}, consists of distinct phases. Initially, the Supervisor orchestrated the code generation phase, sequentially instructing the Coding Agent to produce scripts for the model architecture and training process, and finally, evaluation. Upon receiving confirmation and file paths for each script, the Supervisor would delegate its execution to the Execution Agent. This process ran smoothly until the evaluation stage, where the Execution Agent reported a failure, returning an error log indicating a `FileNotFoundError'. This triggered the system's automated debugging loop. The Supervisor, recognizing the error state, invoked the specialized Tuning Agent, providing it with the failed script and the specific error message through dynamically generated prompts. The Tuning Agent correctly diagnosed the issue, an incorrect relative path in the generated code, and applied a patch using the correct absolute path from the central project context. Upon receiving the patched script, the Supervisor re-ran the execution task for the evaluation script, which then completed successfully. This hierarchical process demonstrates a robust, modular approach to automation that mirrors a well-defined software development pipeline, where specialized roles and a clear chain of command enable reliable error handling and recovery.

\begin{figure}[H]
    \centering
    \includegraphics[width=0.65\textwidth]{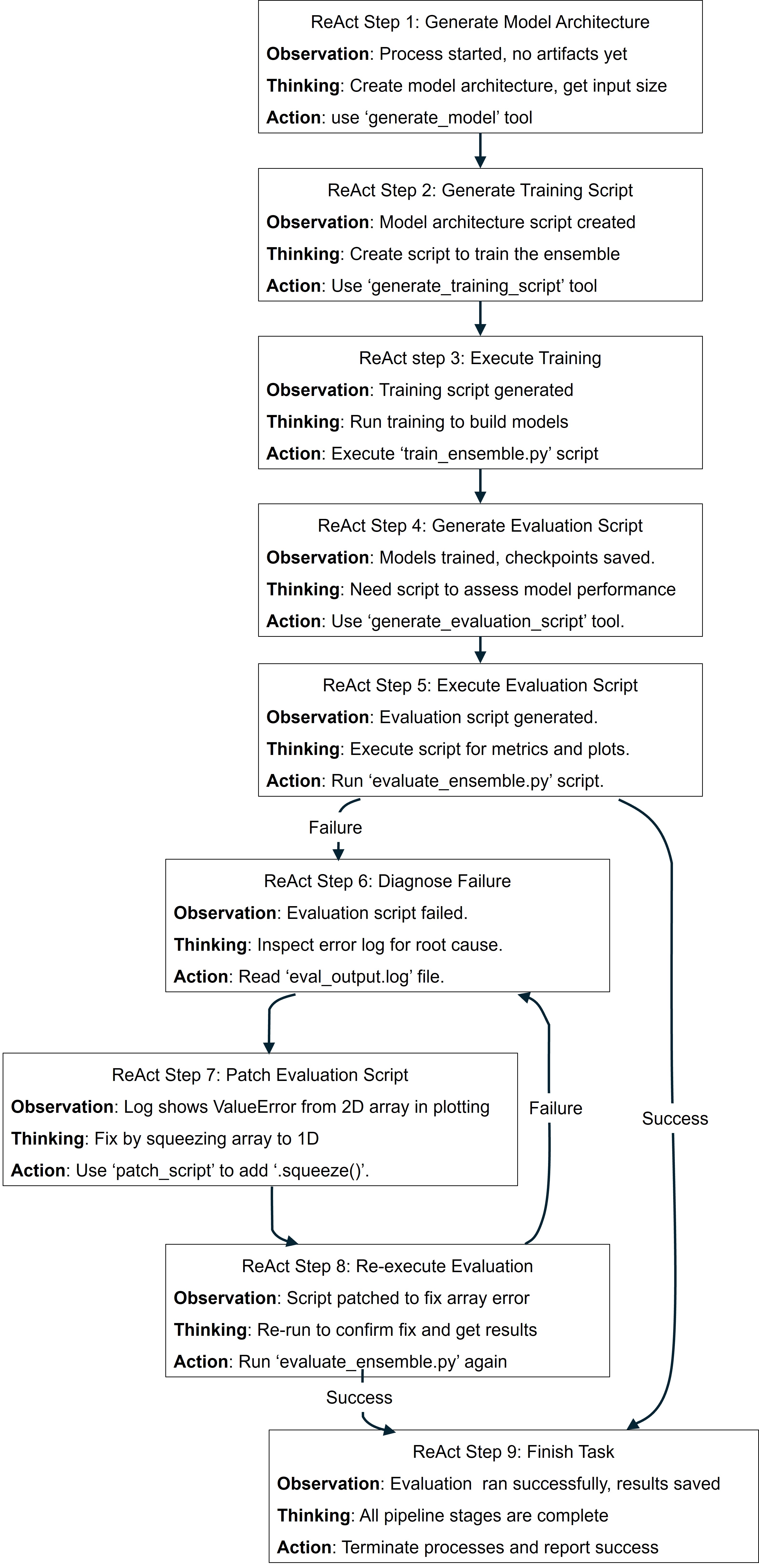}
    \caption{Workflow of the ReAct-agent system illustrating the handling of a code execution failure, and successful re-execution.}
    \label{fig:react_flow}
\end{figure}

On the other hand, the ReAct agent employed a more organic and iterative process guided by its internal thought-action-observation cycle. As demonstrated in Figure \ref{fig:react_flow}, the agent began by reasoning that its first step was to define the model architecture (\textit{Thought}), which led it to call the \textit{generate\_model} tool (\textit{Action}). The successful creation of the model script was fed back as an \textit{Observation}, informing its next \textit{Thought}: to create a training script. This cycle continued seamlessly through model training. The critical test of its adaptability came when it attempted to execute the evaluation script. The \textit{Observation} was a failure, accompanied by an error log. The agent's subsequent \textit{Thought} was to diagnose the failure by reading the log file. From the log, it observed a \textit{ValueError}, correctly reasoning that a 2D data array was being passed to a Matplotlib function that expected a 1D array. This diagnosis led to a new corrective \textit{Action}: patching its own generated code to \textit{.squeeze()} the array into the correct dimension. After confirming the patch was applied, the agent re-attempted the execution, which succeeded. This workflow highlights the ReAct framework's capacity for dynamic, context-aware problem-solving. The agent behaves like a single researcher, iteratively developing a solution, diagnosing unexpected issues based on evidence, and dynamically modifying its own plan and artifacts to overcome them.

These two workflows underscore a key trade-off in agentic design. The multi-agent system offers robustness through specialization and structure, resembling a well-defined engineering pipeline. Its success is based on the clear definition of roles of each agents and the Supervisor's ability to orchestrate them. The ReAct agent, on the other hand, provides flexibility and adaptiveness. Its strength lies in its ability to reason about unforeseen circumstances and formulate solutions at runtime, without relying on pre-defined error handlers for specific failure modes. Both paradigms proved effective in this study, confirming the viability of LLM agents for automating complex scientific analysis and offering distinct models for future development in this domain.

\subsection{Performance Evaluation of LLM Agents-developed Models}

In addition to examining workflow robustness, the predictive performance of models developed by the two LLM agent systems was rigorously evaluated against a human-expert baseline. This evaluation focused specifically on predictive accuracy and the quality of UQ.

\begin{figure}[H]
    \includegraphics[width=1.0\textwidth]{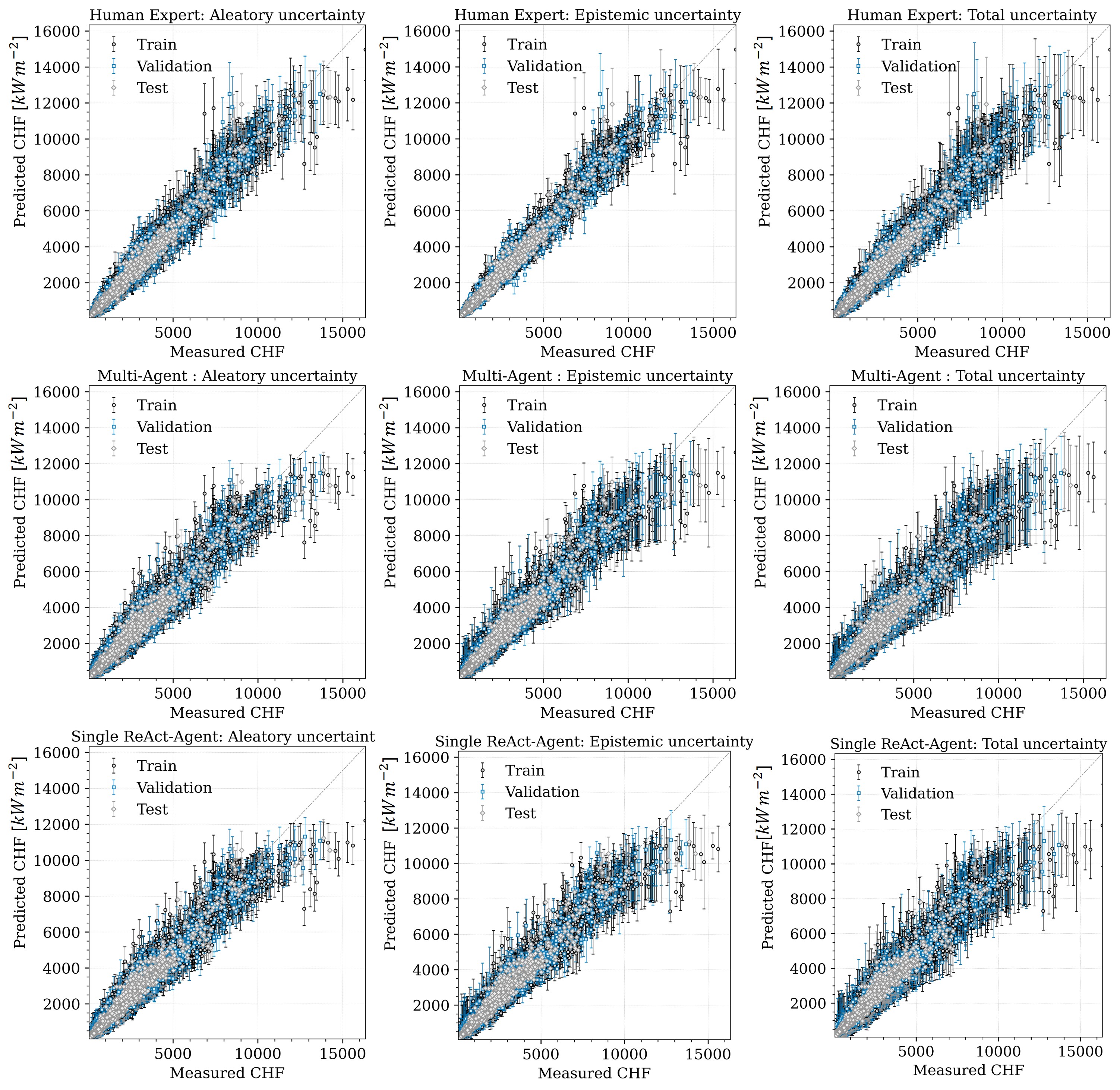}
    \caption{Comparison of deep ensemble predictions and uncertainty quantification for Human Expert, Multi-Agent, and Single ReAct-Agent models against measured CHF data.}
    \label{fig:Pred_compare}
\end{figure}

Figure~\ref{fig:Pred_compare} presents a comprehensive comparison of the predictive performance and UQ for the three developed models: the human-expert baseline, the multi-agent system, and the single ReAct-agent. A qualitative analysis of the plots reveals that all three models demonstrate strong predictive accuracy, with the majority of predictions for the training, validation, and test datasets clustering tightly around the parity line ($y=x$). Additionally, across all three models, the epistemic uncertainty (middle column) is observed to be of a comparable magnitude to the aleatory uncertainty (left column). Aleatory uncertainty represents the inherent, irreducible statistical noise present in the experimental CHF measurements themselves and sets a natural limit on predictive certainty. While epistemic reflects model's own lack of knowledge, often due to data sparsity. The similar scale between these two types of uncertainty suggests both data noise and data sparsity contributes to the overall deep ensemble uncertainty.

Generally speaking, the human expert developed model based on the Bayesian optimized deep ensemble shows a slightly better performance, characterized by a marginally tighter clustering of points and smaller overall uncertainty bounds compared to the two agent-developed models. Quantitatively, the trained Bayesian optimized deep ensemble model has testing data RMSE with 228.9$kW/m^2$; while for Multi-agent developed model and ReAct-agent developed model, this values are 230.2$kW/m^2$ and 234.4$kW/m^2$, respectively. This marginal improvement, however, should be viewed in the context of development overhead: the expert model required time-intensive Bayesian optimization and Gaussian-process modeling, whereas the agents reached similar performance with substantially reduced human effort. Because the two agent models performed similarly with overlapping uncertainty profiles, we report detailed downstream analyses for a single agent model to avoid redundancy. Specifically, the subsequent analysis present results for the Multi-Agent System only, and referred to as the \textbf{LLM agent}.

\begin{figure}[H]
    \includegraphics[width=1.0\textwidth]{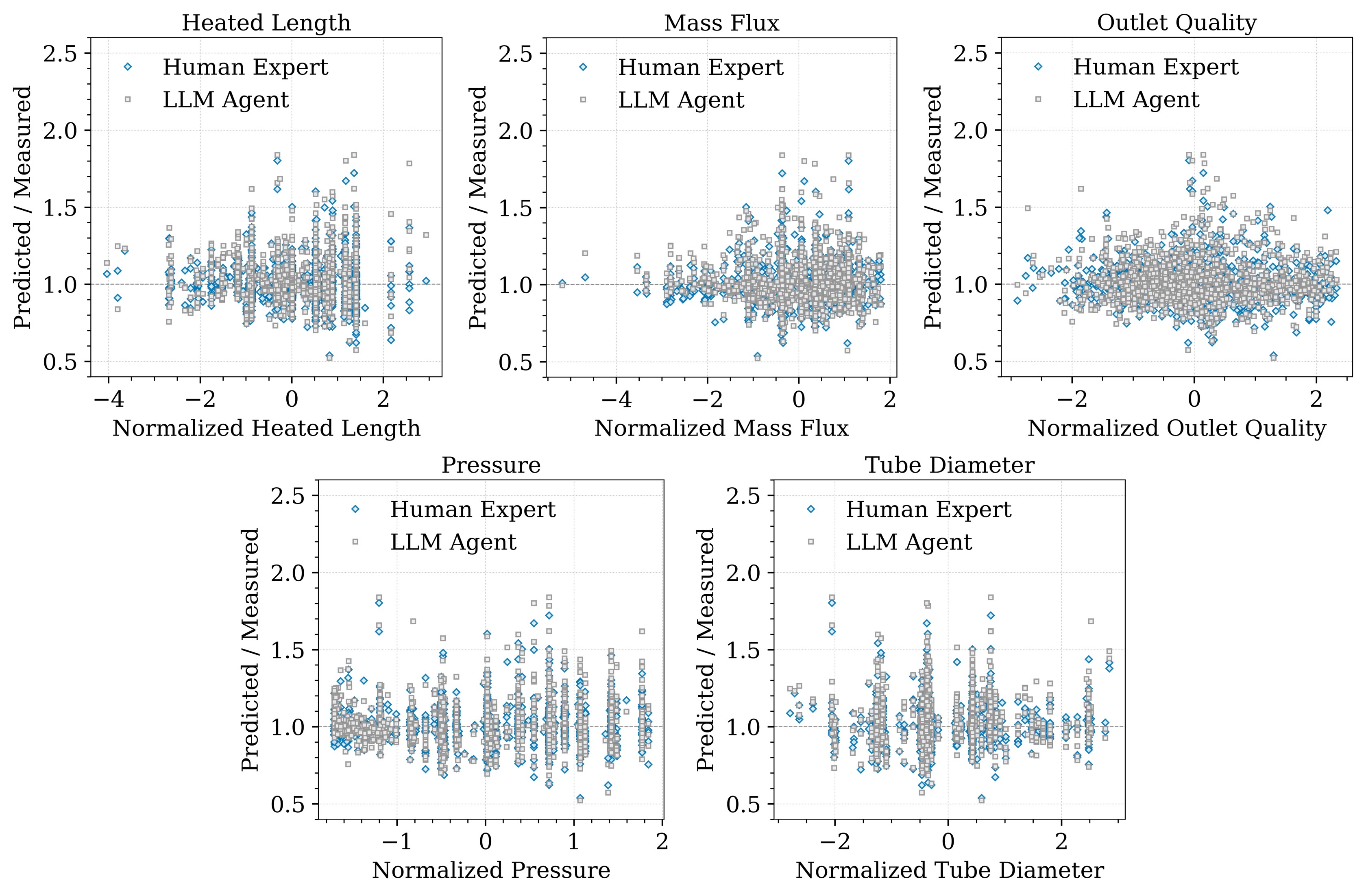}
    \caption{Scatter plots for the test dataset, showing predicted-to-measured CHF ratios as functions of the normalized input features, comparing Human Expert and LLM Agent (Multi-Agent System) models.}
    \label{fig:ratio_compare}
\end{figure}

To further investigate how different features affect model performance, the ratio of prediction to measurement is plotted against each basic input feature, as depicted in Figure~\ref{fig:ratio_compare}. It can be found that both models demonstrate that most prediction ratios fall within a reasonable range (0.5 to 2.0). They also share similar patterns of error distribution; for instance, greater deviations from the ideal ratio of 1.0 are evident at lower values of Mass Flux and higher values of Outlet Quality. A more important observation is the good consistency between the LLM Agent and the Human Expert models. For every input feature, the distribution of prediction ratios from the LLM agent almost perfectly overlaps with that of the human-developed baseline. This indicates that the agent-built model has learned not only the overall predictive trends but also the same subtle biases and performance characteristics as the expert-tuned model.

\begin{figure}[H]
    \includegraphics[width=1.0\textwidth]{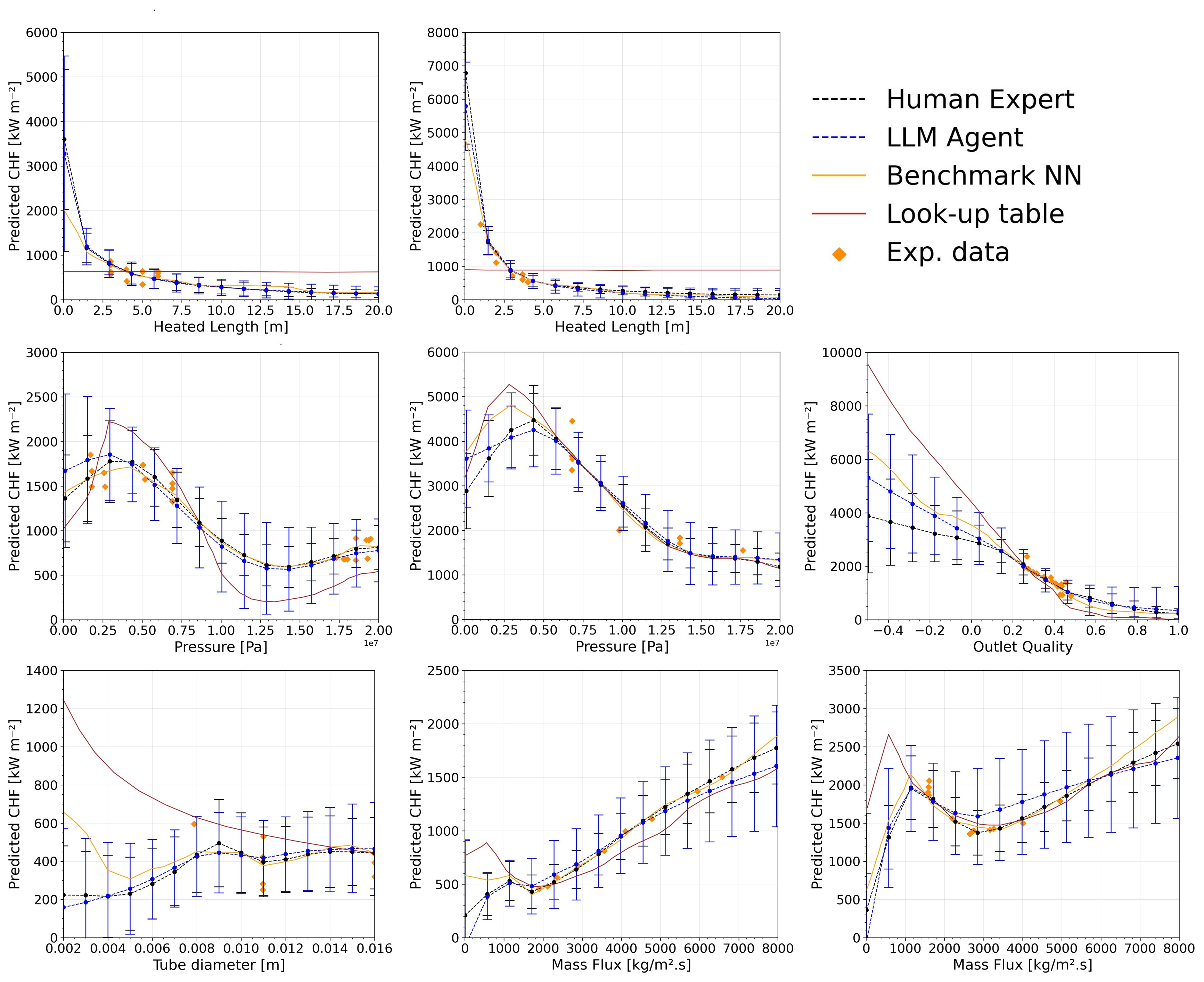}
    \caption{Comprehensive performance evaluation across eight blind datasets, comparing Human Expert, LLM Agent (Multi-Agent System), and look-up table predictions with experimental data, including total uncertainty estimates.}
    \label{fig:sliced_pred}
\end{figure}

To rigorously assess model performance on out-of-distribution data, we evaluated the models using eight ``slice'' datasets from the blind test set as summarizes in Table~\ref{tab:slice_datasets}. Each slice data only has one varying input while holding others nearly constant. The prediction results along with the total uncertainty are depicted in Figure~\ref{fig:sliced_pred}. A major observation is the strong consistency between the LLM agent model and the human-expert baseline across all slices, including challenging regions. In sparsely sampled regimes, e.g., small tube diameter ($<0.006$ m) or high outlet quality ($>0.4$)—both models appropriately increased predictive uncertainty, reflecting heightened epistemic uncertainty. In slices where reference models (e.g., a benchmark neural network or the 2006 CHF lookup table) diverged, the human-expert uncertainty envelope generally encompassed the alternatives, and the LLM agent’s mean closely tracked the expert’s predictions. Notably, the LLM agent model significantly outperformed the CHF lookup table across all slices, providing lower errors and more physically consistent trends. In cases with substantial experimental scatter (e.g., heated length and outlet quality), both the agent and expert models produced smooth trends with appropriately larger uncertainty bounds, capturing the inherent aleatory variability in the data.

\begin{figure}[H]
    \includegraphics[width=1.0\textwidth]{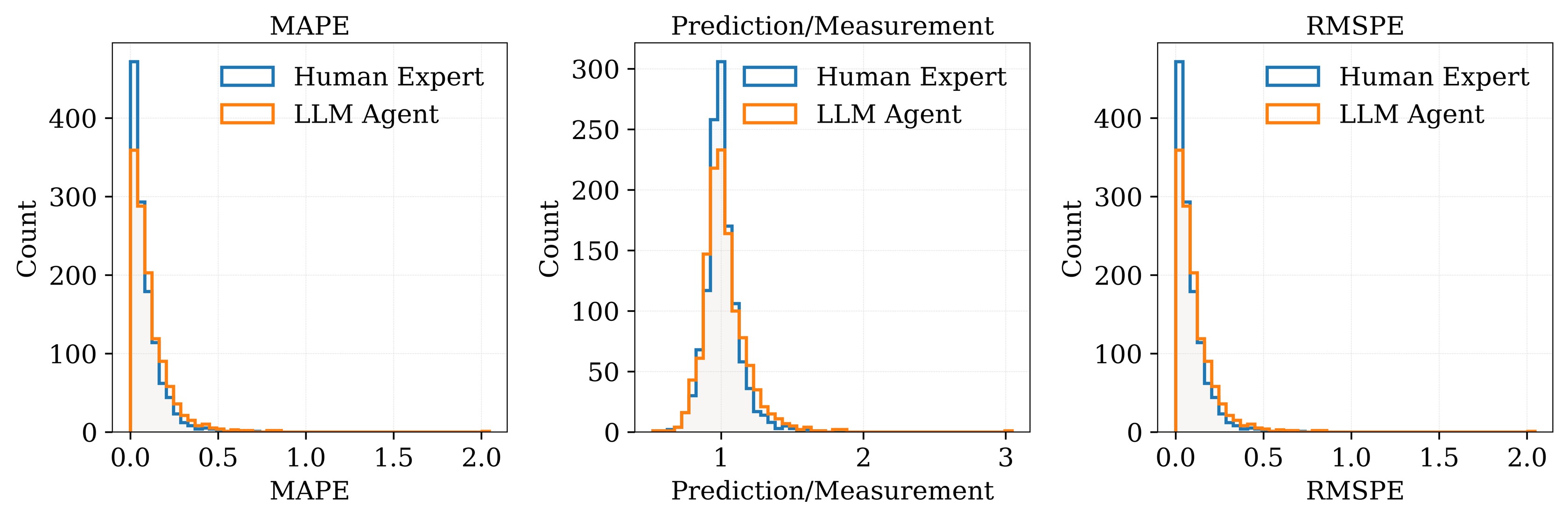}
    \caption{Distribution histograms of error metrics for human expert model and the LLM agent ((Multi-Agent System) developed model.}
    \label{fig:hist_compare}
\end{figure}

To provide a more granular comparison of model performance, Figure~\ref{fig:hist_compare} displays histograms of the error distributions for the Human Expert and LLM Agent models. The comparison is based on three key metrics: the Mean Absolute Percentage Error (MAPE), the Prediction-to-Measurement ratio, and the Root Mean Square Percentage Error (RMSPE).

These metrics are defined as follows. For a set of $n$ samples, where $y_i$ is the true measured value and $\hat{y}_i$ is the model's prediction, the percentage errors are given by:

\begin{equation}
    \text{MAPE} = \frac{100\%}{n} \sum_{i=1}^{n} \left| \frac{y_i - \hat{y}_i}{y_i} \right|
\end{equation}

\begin{equation}
    \text{RMSPE} = \sqrt{\frac{1}{n} \sum_{i=1}^{n} \left( \frac{y_i - \hat{y}_i}{y_i} \right)^2} \times 100\%
\end{equation}

MAPE provides an intuitive measure of the average error magnitude, while RMSPE is more sensitive to large errors due to the squaring term. The Prediction-to-Measurement ratio ($\hat{y}_i/y_i$) is used to assess predictive bias, where a distribution tightly centered at 1.0 is ideal.

The histograms reveal a strong qualitative agreement in the predictive behavior of the two models. The overall shapes of the error distributions for the LLM agent model closely mirror those of the human-developed model, with both being right-skewed and concentrated near the ideal values (0 for percentage errors, 1 for the ratio). This consistency indicates that the agent-developed model successfully captured the fundamental characteristics of the expert-tuned baseline. On the other hand, a subtle performance difference appears in the MAPE and RMSPE plots. The histogram for the human-developed model has a slightly higher peak near zero and a thinner tail compared to the agent's model. This suggests that the human-developed model yielded a higher frequency of very accurate predictions and fewer large errors. Similarly, its Prediction-to-Measurement ratio is more tightly concentrated around 1.0. While this confirms a marginal performance advantage for the human-developed model, the general consistency in the overall error distributions underscores the LLM agent's capability to generate a robust, high-fidelity model.

In summary, these results provide strong evidence that LLM-based agents can autonomously produce data-driven models for complex engineering problems that are on par with state-of-the-art, human-developed counterparts. The agents succeeded not only in achieving high predictive accuracy but also in delivering reliable and physically meaningful UQ, a critical requirement for trustworthy deployment in safety-critical applications. This was all accomplished with minimal human intervention, showcasing the transformative potential of these systems to automate expert-level scientific modeling.

\section{Conclusions}
\label{sec:conclusions}
In this paper, we demonstrate that LLM agents can automate end-to-end data-driven modeling for engineering applications while delivering predictive accuracy and UQ comparable to human expert developed deep ensembles. On the OECD/NEA critical heat flux (CHF) benchmark, both the coordinated multi-agent system and the single ReAct (reasoning and act)-based agent completed the full workflow—from data ingestion through model construction, training, evaluation, and uncertainty analysis—with minimal human intervention. The obtained models significantly outperform the currently widely used CHF lookup table in all blind test cases. Moreover, in addition to point estimates, the agents produced calibrated aleatory and epistemic uncertainty, enabling risk-aware interpretation of results that is essential for risk-informed engineering decision-making.

A central finding is that the two agentic designs succeed through different operational behaviors. The multi-agent framework, organized around a supervisor that orchestrates specialized coding, tuning, and execution roles, tended to be more reliable and computationally efficient across repeated trials. Its division of labor, explicit state management, and targeted self-correction loop yielded consistent runs and lower token usage, making it well suited to structured, throughput-oriented pipelines. By contrast, the single ReAct agent excelled at adaptive problem solving: by interleaving reasoning with tool use, it diagnosed unexpected errors, patched code, and resumed execution without bespoke handlers. This flexibility came with higher variability and resource consumption. Taken together, these results indicate that architecture should be chosen to match operational priorities—multi-agent pipelines for robustness and efficiency; single-agent ReAct for exploratory settings that benefit from dynamic self-repair.

Despite these positive outcomes, we also encountered limitations that point to areas for improvement. One limitation is the current dependency on prompt quality and a bit of human steering; truly hands-free operation required careful prompt engineering and occasionally intervening hints. As LLMs and agent frameworks evolve, reducing this dependency will be important for broader adoption. We also noted that the agents are based on pre-trained LLMs that may lack domain-specific knowledge to implement physical constraints to the model it developed \cite{Abulawi_2025}. Therefore, ensuring robustness and incorporating domain knowledge (when available) remain necessary steps. One approach that will be explored is through the retrieval-augmented-generation to integrate LLM with a vectorized knowledge database \cite{gokdemir2025hiperrag}.

Looking ahead, capability and reliability can be increased by integrating richer tool sets and codified domain knowledge into the agent loop. Tight coupling to simulation codes and structured databases would allow agents not only to analyze data but also to design and run targeted computational experiments, retrieve prior results, and iteratively refine hypotheses, closing the loop between modeling and evidence acquisition \cite{lim2025predicting, lim2025ai}. Combined with improved planning, memory, and checkpointing, these enhancements would expand agent autonomy while preserving traceability.

In sum, LLM agents are a viable path to expert-level, end-to-end modeling in engineering. The multi-agent and ReAct designs provide complementary strengths—robust efficiency versus adaptive flexibility. With deeper tool and knowledge integration and broader validation, such agents can become dependable collaborators in data-rich engineering workflows.

\section*{Data Availability}
The CHF data can be available upon request to the OECD/NEA benchmark activity organizer.

The code for the LLM agents will be available on github after the manuscript is accepted.

\section*{Acknowledgements}

This work is supported by the US Department of Energy Office of Nuclear Energy Distinguished Early Career Program under contract number DE-NE0009468.

\bibliographystyle{elsarticle-num}   

\bibliography{refs} 

\end{document}